\documentclass{article}

     \usepackage[final]{neurips_2020_ml4ps}

\usepackage[utf8]{inputenc} 
\usepackage[T1]{fontenc}    
\usepackage{hyperref}       
\usepackage{url}            
\usepackage{booktabs}       
\usepackage{amsfonts}       
\usepackage{nicefrac}       
\usepackage{microtype}      
\usepackage{graphicx}       

\title{Adversarially trained LSTMs on reduced order models of urban air pollution simulations}

\author{%
    César Quilodrán-Casas\thanks{Corresponding author} \\
  Data Science Institute\\
  Imperial College London\\
  \texttt{cesar.quilodran-casas13@imperial.ac.uk} \\
  \And
  Rossella Arcucci \\
    Data Science Institute\\
    Imperial College London\\
  \texttt{r.arcucci@imperial.ac.uk}
  \And 
  Christopher Pain \\
Department of Earth Science \& Engineering\\
  Imperial College London\\
  \texttt{c.pain@imperial.ac.uk}
  \And
  Yike Guo \\
    Data Science Institute\\
  Imperial College London\\
  \texttt{y.guo@imperial.ac.uk}
}

\begin{document}

\maketitle

\begin{abstract}

This paper presents an approach to improve computational fluid dynamics simulations forecasts of air pollution using deep learning. Our method, which integrates Principal Components Analysis (PCA) and adversarial training, is a way to improve the forecast skill of reduced order models obtained from the original model solution. Once the reduced order model (ROM) is obtained via PCA, a Long Short-Term Memory network (LSTM) is adversarially trained on the ROM to make forecasts. Once trained, the adversarially trained LSTM outperforms a LSTM trained in a classical way. The study area is in London, including velocities and a concentration tracer that replicates a busy traffic junction. This adversarially trained LSTM-based approach is used on the ROM in order to produce faster forecasts of the air pollution tracer.

\end{abstract}

\section{Introduction}

Given the amount of data in Computational Fluid Dynamics (CFD) simulations, data-driven approaches can be seen as attractive solutions to produce reduced order models (ROMs). Furthermore, forecasts produced by these surrogates can be obtained at a fraction of the cost of the original CFD model solution when used together with a ROM. Recurrent neural networks (RNN) have been used to model and predict temporal dependencies between inputs and outputs of ROMs.

Non-intrusive ROMs and RNNs have been used together in previous studies, e.g. \citep{casas2020reduced, reddy2019reduced}. The surrogate forecast systems can easily reproduce a time-step in the future accurately, when data from the original training is available \citep{wu2020data}. However, the problem of these surrogates relies on the error propagation either produced by the ROM or the forecast system, especially when the forecasts wander off the training data. When the predicted output is used as an input for the prediction of the subsequent time sequence, the results can diverge quickly when encountering out-of-distribution data. In real-time applications, like urban air pollution, fast forecasts are urgently needed and an accurate system that can provide reliable forecasts is extremely useful.

A way to obtain reliable forecasts comes from adversarial losses via adversarial training. Generative adversarial networks (GANs) \citep{goodfellow2014generative} are a class of unsupervised machine learning algorithms. In essence, a GAN learns a generative model with the guidance of a discriminator model which is trained jointly, and the loss function provided by the discriminator is referred to as an adversarial loss. However, the idea of adversarial losses can also be applied to supervised scenarios and have advanced the state of the art in many fields over the past years \citep{dong2019towards, wang2019improving}. Additionally, robustness
may be achieved by detecting and rejecting adversarial examples by using adversarial training \citep{shafahi2019adversarial, meng2017magnet}. Data-driven modelling of nonlinear fluid flows incorporating adversarial networks have been successfully being studied previously \citep{cheng2020data, xie2018tempogan}.

This extended abstract applies adversarial training to a LSTM network, based on a ROM of an urban air pollution simulation in an unstructured mesh. The robustness added by the adversarial training allows us to reduce the divergence of the forecast prediction over time, with similar execution times than a LSTM non-adversarially trained.
\section{Methods}

The methodology applies an adversarial training to a supervised ROM-based LSTM, in order to obtain reliable fast forecasts. The non-intrusive ROM is obtained via PCA \citep{arcucci2019domain}. Given that the CFD simulation used in this study contains $O(10^{5})$ dimensions and it is in an unstructured mesh, there are two issues that arise if fast forecasts are needed. 

Firstly, a simulation of this size requires the use of a supercomputer, due to processing power and memory constraint, in order to produce the next time-step. One way to reduce the dimensionality is using PCA, which decomposes the data into linear basis functions that describe the original problem data. The caveat of using PCA is that the dimension reduction comes from truncating the Principal Components (PC), and by doing so the retained variance decreases. However, for the purposes of the forecast only, only a truncated PCA dimension reduction is used here.

Secondly, a traditional 3D Convolutional Neural Network (CNN) is not trivial since the CFD simulation lies on a unstructured mesh rather than on a regular grid \citep{kim2019deep, casas2020urban}.

\begin{figure*}[h!]
  \centering
  \includegraphics[trim = {0cm 1cm 0cm 1cm},clip,width =1\textwidth]{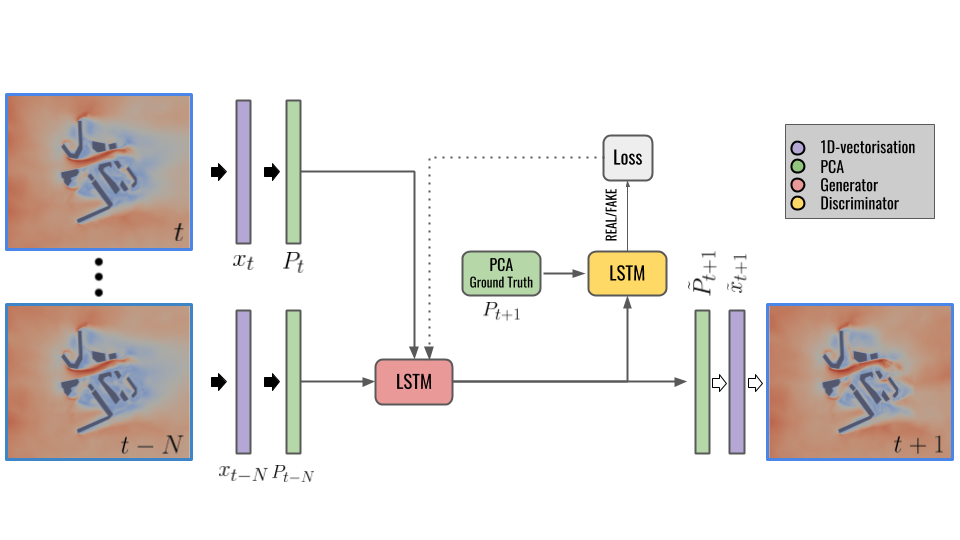}
  \caption{Proposed workflow for adversarial training of LSTM. The dimension-reduced data is used as an input in the LSTM, and its output is judged by the LSTM discriminator. The LSTM takes $N$ previous time-steps to produce a forecast of $t+1$. The vectorisation translates each point of the unstructured mesh onto a 1-dimensional vector $\mathbf{x}$. Then, $\mathbf{P}$ are the principal components of $\mathbf{x}$.}
  \label{fig:workflow}
  \end{figure*}

The workflow is presented in Figure \ref{fig:workflow}.

\subsection{Principal Components Analysis}

As described by \citet{lever2017points} PCA is an unsupervised learning method that simplifies high-dimensional data by transforming it into fewer dimensions. Let 
\begin{equation}
    \mathbf{x}=\left\{ \mathbf{x}_t \right\}_{t=1,\dots,n}
\end{equation}

$\mathbf{x}\in \Re^{n \times m}$, with $n<m$, denotes the matrix of the model vectors at each time step. The PCA consists in decomposing this dataset as:
\begin{eqnarray}
    \mathbf{x} &=& \mathbf{P} \mathbf{\Pi} + \mathbf{\bar{x}} \\
    \mathbf{x}_{\tau} &=& \mathbf{P}_{\tau} \mathbf{\Pi}_{\tau} + \mathbf{\bar{x}}
\end{eqnarray}
%
where $\bf{P} \in \it{\mathbb{\Re}^{n\times n}}$ are the principal components of $\mathbf{x}$; $\bf{\Pi} \in \it{\mathbb{\Re}^{n\times m}}$ are the Empirical Orthogonal Functions; and $\mathbf{\bar{x}}$ is the mean vector of the model. The dimension reduction of the system comes from truncating $\mathbf{P}$ at the first $\tau$ PCs, with $\mathbf{P}_{\tau} \in \Re^{n\times \tau}$.

\subsection{Adversarially trained long short-term memory}
\label{sec:adv}
Once the truncated PCs are obtained, they can be used to train a LSTM \citep{hochreiter1997long} to make predictions. In this paper, a vanilla LSTM network takes $N$ previous time-steps of $\mathbf{P}_{t-N},\dots, \mathbf{P}_{t}$ as input and predicts an approximation of $\mathbf{P}_{t+1}$, named $\mathbf{\tilde{P}}_{t+1}$:

\begin{equation}
f^{LSTM}:\mathbf{P}_{t-N},\dots, \mathbf{P}_{t} \to  \mathbf{\tilde{P}}_{t+1}.
\end{equation}

Our methodology proposes to add adversarial training to the LSTM network is order to further improve the forecasts.

The supervised adversarial training includes a discriminator to distinguish between the real samples of the Principal Components $\mathbf{P}$ and predictions $\mathbf{\tilde{P}}$ produced by  $f^{LSTM}$. The $\mathbf{P}$ are fed to the discriminator as real sequences (ground truth). Let, $D(x, y)$ represent the discriminator function with an input $x$ and a target label $y$ such that, for $x=\mathbf{P}_{t+1}$, $y=1$ and for $x=\mathbf{\tilde{P}}_{t+1}$, $y=0$ . The training of $D$ is based on the minimisation of the binary cross-entropy loss ($\mathcal{L}^{bce}$), using the Nesterov Adam optimizer (Nadam) \citep{dozat2016incorporating}. The loss of $f^{LSTM}$ also includes the mean squared error (mse) between $\mathbf{\tilde{P}}_{t+1}$ and $\mathbf{P}_{t+1}$.

The adversarial losses $\mathcal{L}^{adv}$ for $D$ and $f^{LSTM}$ are then defined as:
\begin{eqnarray}
    \mathcal{L}^{adv}_{D}(\mathbf{P}_{t+1}) &=& \mathcal{L}^{bce}(D(\mathbf{P}_{t+1}, 1)) + \mathcal{L}^{bce}(D(f^{LSTM}(\mathbf{P}_{t-N},\dots, \mathbf{P}_{t}), 0))\\
    \mathcal{L}^{adv}_{f^{LSTM}}(\mathbf{P}) &=& \mathcal{L}^{bce}(D(f^{LSTM}(\mathbf{P}_{t-N},\dots, \mathbf{P}_{t}), 1)) + \mathcal{L}^{mse}(f^{LSTM}(\mathbf{P}_{t-N},\dots, \mathbf{P}_{t}))
\end{eqnarray}
%



\section{Study area, model data and results}
The computational fluid dynamics (CFD) simulations were carried out using Fluidity \citep{davies2011fluidity} (\url{http://fluidityproject.github.io/}). The study area is a 3D realistic representation of a part of South London. The dispersion of the pollution is described by the classic advection-diffusion equation (eq.~(\ref{Eq:TracerEq})).
\begin{equation}\label{Eq:TracerEq}
    \frac{\partial c}{\partial t}+\nabla.(\mathbf{u}c)=\nabla.\left(\overline{\overline{\kappa}}\nabla c\right)+F
\end{equation}
where $\overline{\overline{\kappa}}$ is the diffusivity tensor (m\textsuperscript{2}/s), $c$ is the concentration and $F$ represents the source terms (kg/m\textsuperscript{3}/s). The passive tracer is the transport of a scalar field representing a point source mimicking pollution in a traffic congested junction whose advection and diffusion depend on velocity, pressure and density. 

The 3D case is composed of an unstructured mesh including $m = 100,040$ nodes per dimension and $n = 1500$ time-steps. The wind profile of the atmospheric boundary layer is represented by a log-profile velocity. The top and the sides of the model domain have a perfect slip boundary condition, while the facades of the buildings and the bottom of the model domain have a no-slip boundary. The pollution background is modelled as a sinusoidal function. This background pollution mimics waves of pollution in an urban environment. 

\section{Experiments and results}

Four experiments were set up to assess the improved forecast of the adversarially trained LSTM. A PCA was applied to two output fields from the CFD simulation: Tracer (1-dimensional, unitless), and Velocity (3-dimensional, $m s^{-1}$). To each of these set of PCs, a truncation of 64 and 128 PCs was applied which explains over 90\% of variance in each case. The truncation to 64 PCs and 128 PCs reduces the size of the dataset by 4  and 3 orders of magnitude, respectively. Only 90\% of the data is used for training, and 10\% is used for validation.

Firstly, a grid search for optimal hyperparameters was set up for each of these experiments, including dropouts, hidden nodes in the LSTM, batch sizes, activation function of the output layer, and lags. Once the optimal set of hyperparameters for each experiment is found, the trained LSTM (named $LSTM^{classic}$) is saved. 

Secondly, the adversarially trained LSTM (named $LSTM^{adv}$) is setup with the same optimal hyperparameters found for each combination. The adversarial training includes a discriminator $D$. The discriminator $D$ is a LSTM that takes $\mathbf{\tilde{P}}$ and outputs between 0 and 1 (sigmoid function) as described in Section \ref{sec:adv}. 
The hyperparameters used to train the LSTM networks for each experiment are summarised in table \ref{table:hyperparameters}. The adversarial training adds the same mirrored architecture as a discriminator $D$ with an output dense layer of size 1 with a sigmoid activation function, and binary cross-entropy as loss. Both $LSTM^{classic}$ and $LSTM^{adv}$ were trained for 5000 epochs with a piece-wise mean squared error loss.

\begin{table*}[ht]
\centering
  \caption{Hyperparameters for each experiment. AF is the activation function in the output layer, and Time-lag is the number of time-steps used to predict the following one.}
  \label{table:hyperparameters}
  \begin{tabular}{|c|c|c|c|c|c|}
    \hline
    Experiments & Batch size & Hidden nodes LSTM & Dropout & AF & Time-lag\\
    \hline
    Tracer $\tau$ = 64 & 32 & 256 & 0.5 & ReLU & 2\\
    Tracer $\tau$ = 128 & 32 & 256 & 0.3 & Sigmoid & 2\\
        Velocity $\tau$ = 64 & 32 & 128 & 0.5 & ReLU & 2\\
    Velocity $\tau$ = 128 & 32 & 256 & 0.5 & Sigmoid & 2\\
    \hline
    \end{tabular}
\end{table*}

\begin{figure*}[t]
  \centering
  \includegraphics[trim = {0cm 0cm 0cm 0cm},clip,width = 1\textwidth]{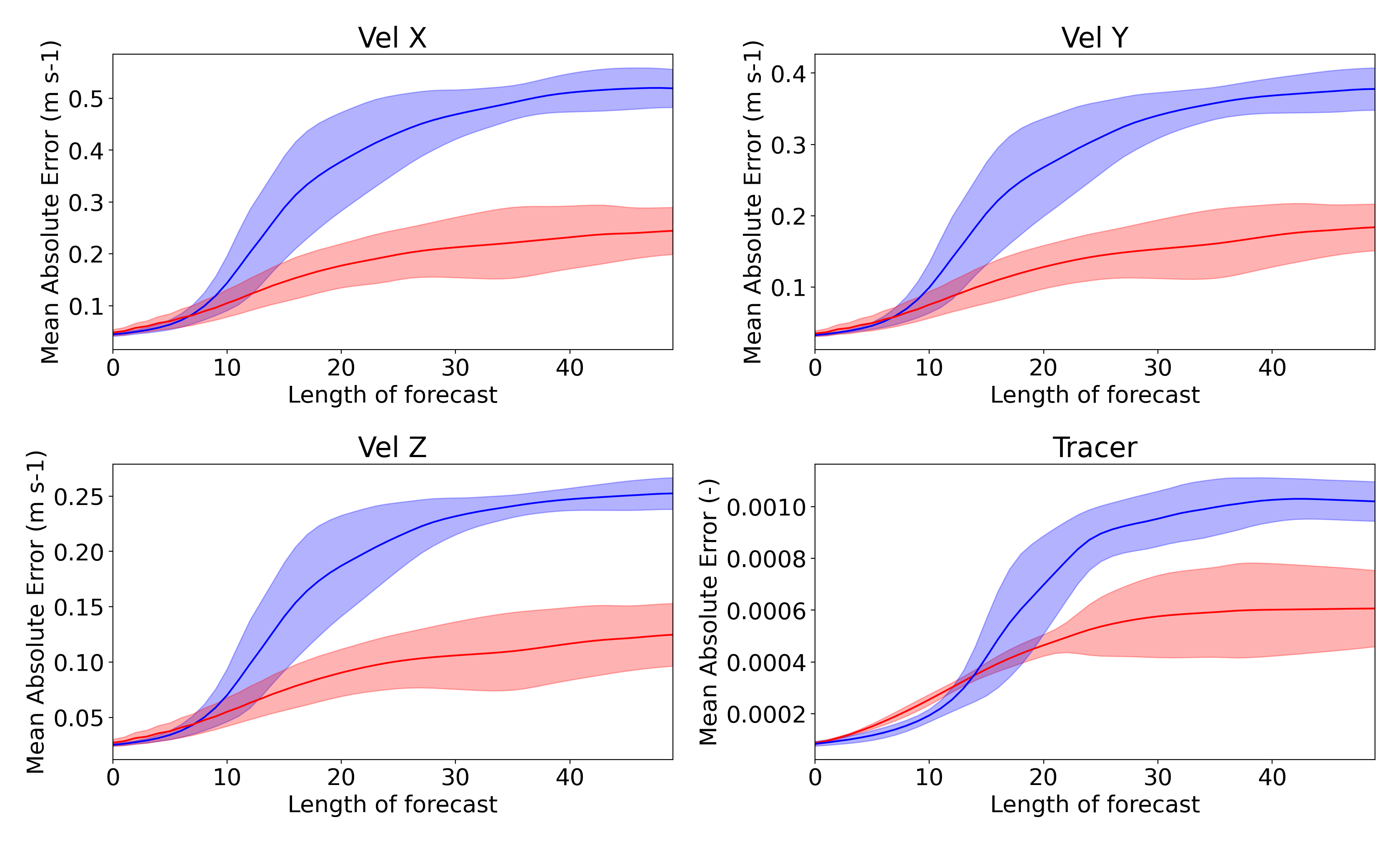}
  \caption{Comparison of $LSTM^{classic}$ (blue) and $LSTM^{adv}$ (red). The shaded areas show an ensemble of errors of 50 time-step long forecast from different starting points from $t=350$ to $t=400$ (within training data) with 128 PCs. The solid line is the mean and the shaded area is one standard deviation from the mean.}
  \label{fig:ensembleforecast}
\end{figure*}

Figure \ref{fig:ensembleforecast} presents the error from an ensemble of forecasts, of velocities in X, Y, Z and Tracer, starting from different time-steps. The solid line represents the mean of the error ensemble and the shaded area is the standard deviation from the mean. The red shaded area is $LSTM^{adv}$, while the blue shaded area is $LSTM^{classic}$. The forecasts are created by using previous time-steps from data and producing a forecast. This forecast is subsequently used as an input for the prediction of the next time-step. After 50 iterations, it is very clear that $LSTM^{adv}$ outperforms $LSTM^{classic}$. The forecasts are 4 orders of magnitude faster than the CFD simulation. Table \ref{table:errorreduction} shows further results for all the experiments.

\begin{table*}[ht]
\centering
  \caption{Forecast improvement, error reduction in $\%$ of $LSTM^{adv}$ over $LSTM^{classic}$.}
  \label{table:errorreduction}
  \begin{tabular}{|c|c c c c| c c c c|}
    \hline
    & \multicolumn{4}{c|}{Within training data} & \multicolumn{4}{c|}{Within validation data}\\
    \cline{2-9}
    & \multicolumn{3}{c}{Velocity} & Tracer (c) & \multicolumn{3}{c}{Velocity} & Tracer (c)\\
    PCs & X & Y & Z & & X & Y & Z & \\
    \hline
    $\tau$ = 64 & 21.07 & 22.21 & 83.32 & 18.95 & 11.90 & 12.14 & 13.52 & 9.01\\
    $\tau$ = 128 & 51.09 & 51.39 & 49.13 & 56.29 & 8.13 & 8.18 & 10.02 & 6.46\\
    \hline
    \end{tabular}
\end{table*}

\begin{figure}[t]
  \centering
  \includegraphics[trim = {1cm 1cm 1cm 1cm},clip,width = 1\textwidth]{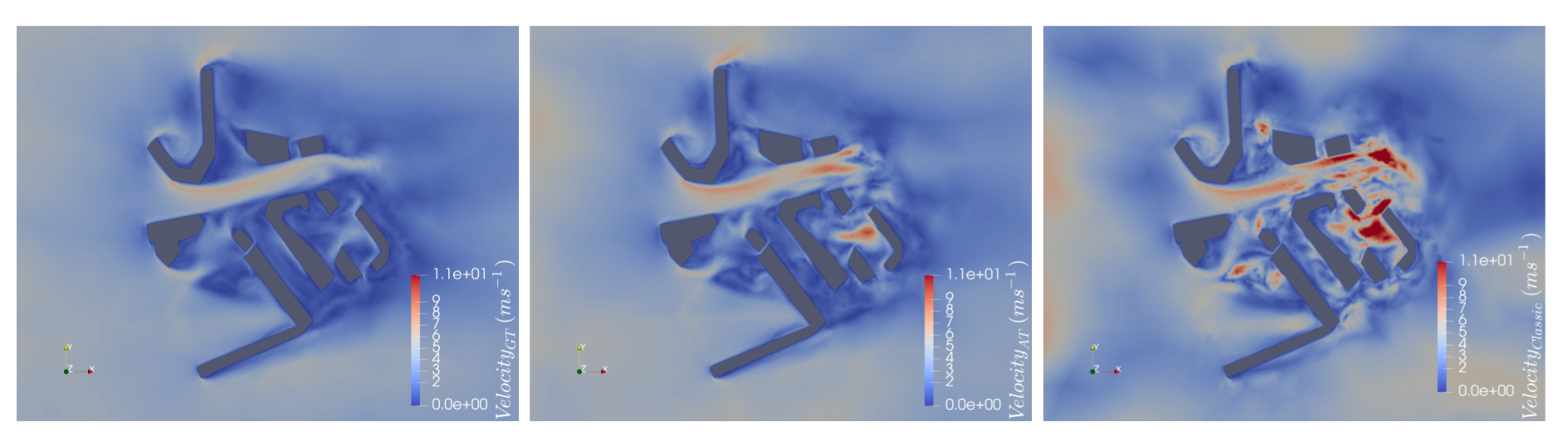}
  \caption{Comparison of forecasted velocity in $m s^{-1}$ (magnitude) fields by $LSTM^{adv}$ and $LSTM^{classic}$ with 64 PCs. This is a 25 time-step forecast starting from $t=350$. Left: Ground truth at $t=375$, Middle: $LSTM^{adv}$ forecasted 25 time-steps from $t=350$, Right: $LSTM^{classic}$ forecasted 25 time-steps from $t=350$.}
  \label{fig:vel_fields}
\end{figure}

Figure \ref{fig:vel_fields} shows the comparison of forecasted magnitude velocity (in $m s^{-1}$) fields of $LSTM^{adv}$ and $LSTM^{classic}$ using 64 PCs from $t=350$. The snapshots show clearly that after 25 time-steps of forecasting, $LSTM^{classic}$ diverges quickly from the underlying model state, while $LSTM^{adv}$ preserve more underlying physics.

\section{Summary and future work}

This paper presented an adversarially trained LSTM that improves the forecast of a LSTM trained in a classical way. This is important when accurate near real-time predictions are needed and not enough data is available. It can be observed that adversarially trained LSTM does not diverge greatly from the data it has learned, given the constraint of the discriminator network.   

The replacement of the CFD solution by these models will speed up the forecast process towards a real-time solution. And, with the application of adversarial training could potentially produce more physically realistic flows. Future work will apply the same methodology to different dimension reduction schemes. Furthermore, this framework is data-agnostic and could be applied to different CFD models where enough data is available.

\clearpage
\section*{Acknowledgements}

This work is supported by the EPSRC Grand Challenge grant ‘Managing Air for Green Inner Cities (MAGIC) EP/N010221/1, the EPSRC Centre for Mathematics of Precision Healthcare EP/N0145291/1 and the EP/T003189/1 Health assessment across biological length scales for personal pollution exposure and its mitigation (INHALE). Thanks to Dr. Laetitia Mottet for the set up of the full model in Fluidity.

\section*{Broader Impact}

In this paper, researchers introduced an adversarially trained LSTM to improve the forecast of reduced order models of urban air pollution. The impact of this research is likely to drive the development of improved urban air pollution forecasts, at a fraction of the computational costs but equally as accurate. Our framework comes from the urgency to create data-driven real-time forecasts from expensive computational fluid dynamics simulations. Furthermore, our framework is data-agnostic and can be applied to different variables or simulations where there is enough available data.

\bibliographystyle{abbrvnat}
\bibliography{references.bib}

\small

\end{document}